\documentclass[conference]{IEEEtran}
\IEEEoverridecommandlockouts
\usepackage{cite}
\usepackage{amsmath,amssymb,amsfonts}
\usepackage{siunitx}

\usepackage{graphicx}
\usepackage{textcomp}
\usepackage{xcolor}
\usepackage[lofdepth,lotdepth]{subfig}

\def\BibTeX{{\rm B\kern-.05em{\sc i\kern-.025em b}\kern-.08em
   T\kern-.1667em\lower.7ex\hbox{E}\kern-.125emX}}

\usepackage{amssymb}
\usepackage{amsfonts}

\usepackage{xcolor}
\usepackage{ulem}
\usepackage[super]{nth}

\usepackage[colorlinks=true]{hyperref}
\usepackage{url}

\usepackage [boxruled, rightnl]{algorithm2e}
\usepackage{algpseudocode}

\newcommand{\mut}{$\mu$T}

\newcommand{\mb}[1]{\mathbf{#1}}

\DeclareMathOperator{\diag}{diag}
\DeclareMathOperator{\dhat}{\hat{d}}

\DeclareMathAlphabet{\pazocal}{OMS}{zplm}{m}{n}
\newcommand{\x}{\textbf{x}}
\newcommand{\z}{\textbf{z}}

\renewcommand{\u}{\textbf{u}}
\newcommand{\Par}{\pazocal{X}}
\newcommand{\Wei}{\pazocal{W}}
\newcommand{\Particles}{\Par_t}
\newcommand{\ParticlesPrev}{\Par_{t-1}}

\newcommand{\ParticlesOver}{\overline{\Par}_t}

\newcommand{\Weights}{\Wei_t}

\newcommand{\WeightsPrev}{\Wei_{t-1}}

\newcommand{\xIT}{\x_t^{(i)}}
\newcommand{\xITPrev}{\x_{t-1}^{(i)}}
\newcommand{\mIT}{m_t^{(i)}}
\newcommand{\mITPrev}{m_{t-1}^{(i)}}

\newcommand{\wIT}{w_t^{(i)}}
\newcommand{\wITPrev}{w_{t-1}^{(i)}}

\newcommand{\motionModelI}{p(\x | \xITPrev, \u_t)}

\begin{document}

\title{Saying goodbyes to rotating your phone: \\ Magnetometer calibration during SLAM 
{\footnotesize 
}
\thanks{\copyright 2024 IEEE. Personal use of this material is permitted. Permission from IEEE must be obtained for all other uses, in any current or future media, including reprinting/republishing this material for advertising or promotional purposes, creating new collective works, for resale or redistribution to servers or lists, or reuse of any copyrighted component of this work in other works. 

Accepted for publication at the \nth{14} International Conference on Indoor Positioning and Indoor Navigation (IPIN 2024).}
}

\author{\IEEEauthorblockN{1\textsuperscript{st} Ilari Vallivaara}
\IEEEauthorblockA{\textit{Visiting Research Fellow} \\
\textit{University of Edinburgh}\\
Edinburgh, UK 
}
\and
\IEEEauthorblockN{2\textsuperscript{nd} Yinhuan Dong}
\IEEEauthorblockA{\textit{School of Engineering} \\
\textit{University of Edinburgh}\\
Edinburgh, UK 
}
\and
\IEEEauthorblockN{3\textsuperscript{rd} Tughrul Arslan}
\IEEEauthorblockA{\textit{School of Engineering} \\
\textit{University of Edinburgh}\\
Edinburgh, UK 
}
}

\maketitle

\begin{abstract}
While Wi-Fi positioning is still more common indoors, using magnetic field features has become widely known and utilized as an alternative or supporting source of information. Magnetometer bias presents significant challenge in magnetic field navigation and SLAM. Traditionally, magnetometers have been calibrated using standard sphere or ellipsoid fitting methods and by requiring manual user procedures, such as rotating a smartphone in a figure-eight shape. This is not always feasible, particularly when the magnetometer is attached to heavy or fast-moving platforms, or when user behavior cannot be reliably controlled. Recent research has proposed using map data for calibration during positioning. This paper takes a step further and verifies that a pre-collected map is not needed; instead, calibration can be done as part of a SLAM process. The presented solution uses a factorized particle filter that factors out calibration in addition to the magnetic field map. The method is validated using smartphone data from a shopping mall and mobile robotics data from an office environment. Results support the claim that magnetometer calibration can be achieved during SLAM with comparable accuracy to manual calibration. Furthermore, the method seems to slightly improve manual calibration when used on top of it, suggesting potential for integrating various calibration approaches.
\end{abstract}


\section{Introduction} \label{section:introduction}
Despite active research, indoor positioning lacks a dominant solution comparable to GPS. In addition to Wi-Fi-based approaches \cite{mendoza2019meta, torres2016providing, gao2017signal}, using the indoor magnetic field (MF) for positioning has gained popularity over the last 15 years \cite{torres2016providing, gao2017signal, haverinen2009global, li2012feasible, angermann2012characterization, solin2016terrain, siebler2023magnetic_robot}.
Beyond positioning, MF can also be used for Simultaneous Localization And Mapping (SLAM), where an agent builds a map while positioning itself on it. Several studies have shown the MF's capacity to correct drift inherent to SLAM \cite{robertson2013simultaneous, kok2018scalable, viset2022extended, pavlasek2023magnetic, vallivaara2011magnetic}. 
The applications of MF-based positioning extend beyond mobile robots and smartphone-carrying humans to include airplanes \cite{lee2020magslam, canciani2020analysis}, marine robots \cite{teixeira2017robust}, and trains \cite{siebler2023simultaneous_train}.

\begin{figure}[ht!]
\centering
\subfloat[][Calibration movement]{
\includegraphics[width=1.0\columnwidth]{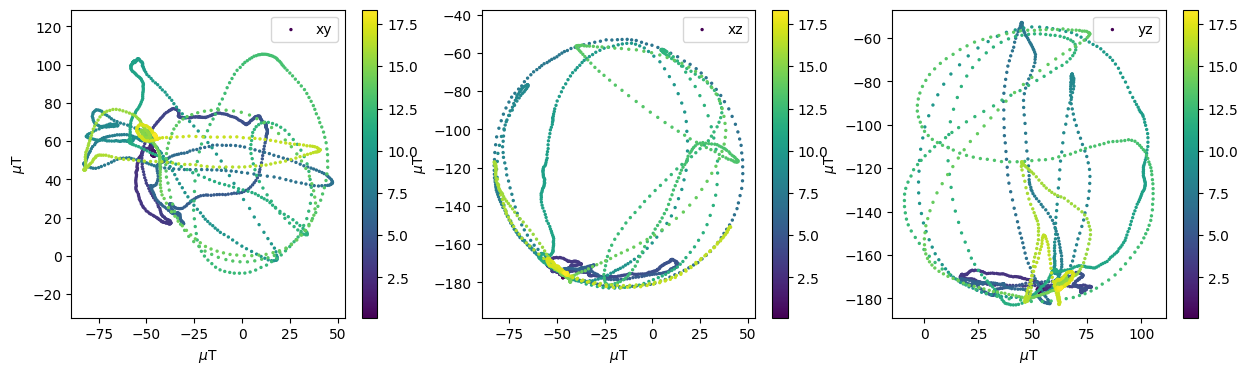}
\label{fig:calib_movement_calib}}
\qquad
\subfloat[Subfigure 3 list of figures text][Natural walk data]{
\includegraphics[width=1.0\columnwidth]{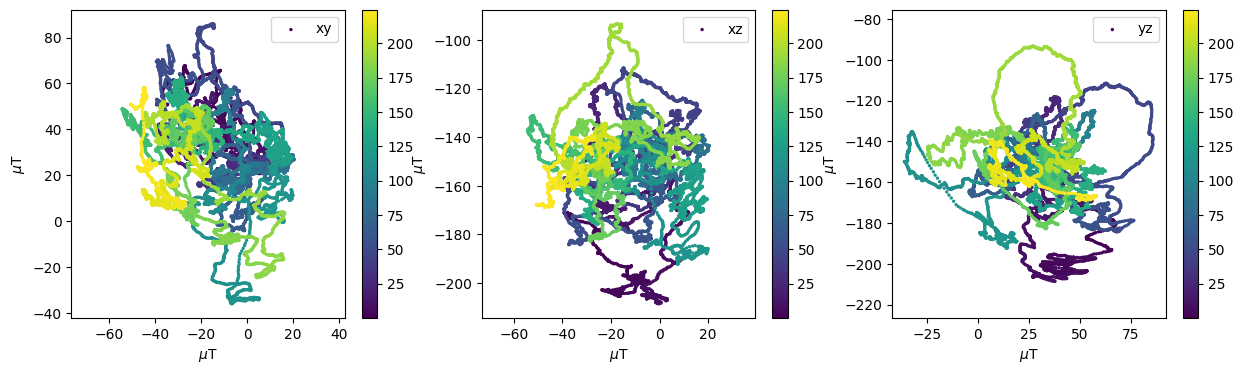}
\label{fig:calib_movement_traj}}
\qquad
\caption{Two-axis projections of the magnetic field signal during a calibration movement (top) and a trajectory in a shopping mall (bottom), illustrated in Fig. \ref{fig:calib_estimation}. The color indicates time in seconds. Bias $\mb{b} \in \mathbb{R}^3$ can be extracted from the calibration movement using sphere fitting methods. This paper demonstrates bias estimation from natural walk data.}
\label{fig:calib_movement}
\end{figure}

A major challenge in MF-based positioning is ensuring proper calibration of often biased magnetometers \cite{gao2017signal, kok2016magnetometer}, as biased MF data affects positioning very negatively. Calibration typically involves manual rotation of the magnetometer in a homogeneous magnetic field and applying sphere or ellipsoid fitting methods to determine calibration parameters \cite{haverinen2009global, solin2016terrain, kok2016magnetometer, angermann2012characterization}.  Figure \ref{fig:calib_movement_calib} illustrates typical data from such a calibration procedure. Smartphone operating systems (OS), like Android, run background sphere fitting methods to provide calibrated readings alongside raw data. However, this uncontrolled calibration can be unpredictable and unstable, often reliable only immediately after manual rotation. 

\begin{figure}[ht]
\centering
\includegraphics[width=1.0\columnwidth]{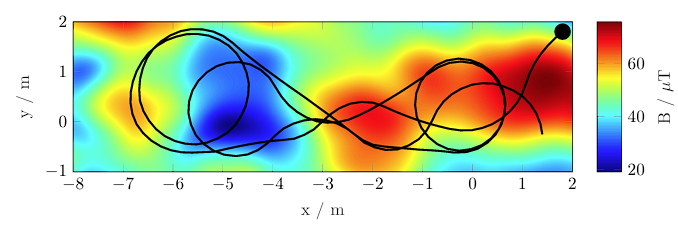}
\caption{ A pre-collected MF map and a mobile robot trajectory from DLR's laboratory from Siebler et al's work on Simultaneous Localization and Calibration (SLAC) \cite{siebler2023magnetic_robot}. The authors are able to estimate the bias during positioning by using a factorized PF and a KF. Permission for reuse granted by IEEE (©2023 IEEE).}
\label{fig:siebler}
\end{figure}

Rotating the magnetometer in place is not always feasible, particularly when mounted on heavy or fast-moving platforms like trains \cite{siebler2023simultaneous_train} or planes \cite{lee2020magslam}. Additionally, reliably controlling the behavior of human agents carrying the sensor can be challenging. Addressing the first set of challenges, Siebler et al. have proposed calibration methods using map data in both 1D train \cite{siebler2023simultaneous_train} and 2D mobile robot scenarios \cite{siebler2023magnetic_robot}. This paper is heavily inspired by their Simultaneous Localization And Calibration (SLAC) approach \cite{siebler2023magnetic_robot} for mobile robotics, which uses a factorized Particle Filter (PF) on a pre-collected MF map (Fig. \ref{fig:siebler}) and estimates calibration for each particle using a Kalman Filter (KF). In contrast to their work, our approach does not rely on a pre-collected MF map.

The concept of \textit{Mapping with known poses} \cite{stachniss2007analyzing} relates to Particle Filter SLAM, where the joint posterior of the map and the poses is factorized into two parts: The poses are estimated using the PF, and the map is built analytically for each particle based on its history and measurements. 

\section{Intuition} \label{section:intuition}
Manual calibration makes bias observable through rotation in a constant field. This is true to a lesser degree if we have fewer data points from a small spatial neighborhood; the main idea of this paper is to find and evaluate those neighborhoods using a particle filter, while estimating the bias with a KF carried by each particle. We build on both the bias and the map factorization and propose \textit{calibration and mapping with known poses}, where each particle carries a bias estimate (KF) in addition to a MF map. Our experiments demonstrate that this approach can simultaneously estimate calibration, construct the map, and position the agent without manual calibration.

\subsection{Calibration with known poses}
Data from a natural walk differs significantly from calibration data  (compare Fig. \ref{fig:calib_movement_calib} and \ref{fig:calib_movement_traj}). However, if we assume knowledge of the exact position for each measurement and group spatially close measurements together, we obtain a set of partial sphere surfaces with internally homogeneous MF. Particularly with sensor orientations, aligning measurements to a common frame allows us to find a bias that best fits these surfaces. Unfortunately, we usually lack direct access to these positions, and the search space is huge. The calibration movement is essentially calibration with \textit{a single known pose}.

\subsection{Calibration and mapping with known poses}
\begin{figure}[t]
\centering
\subfloat[][St James Quarter shopping mall]{
\includegraphics[width=1.0\columnwidth]{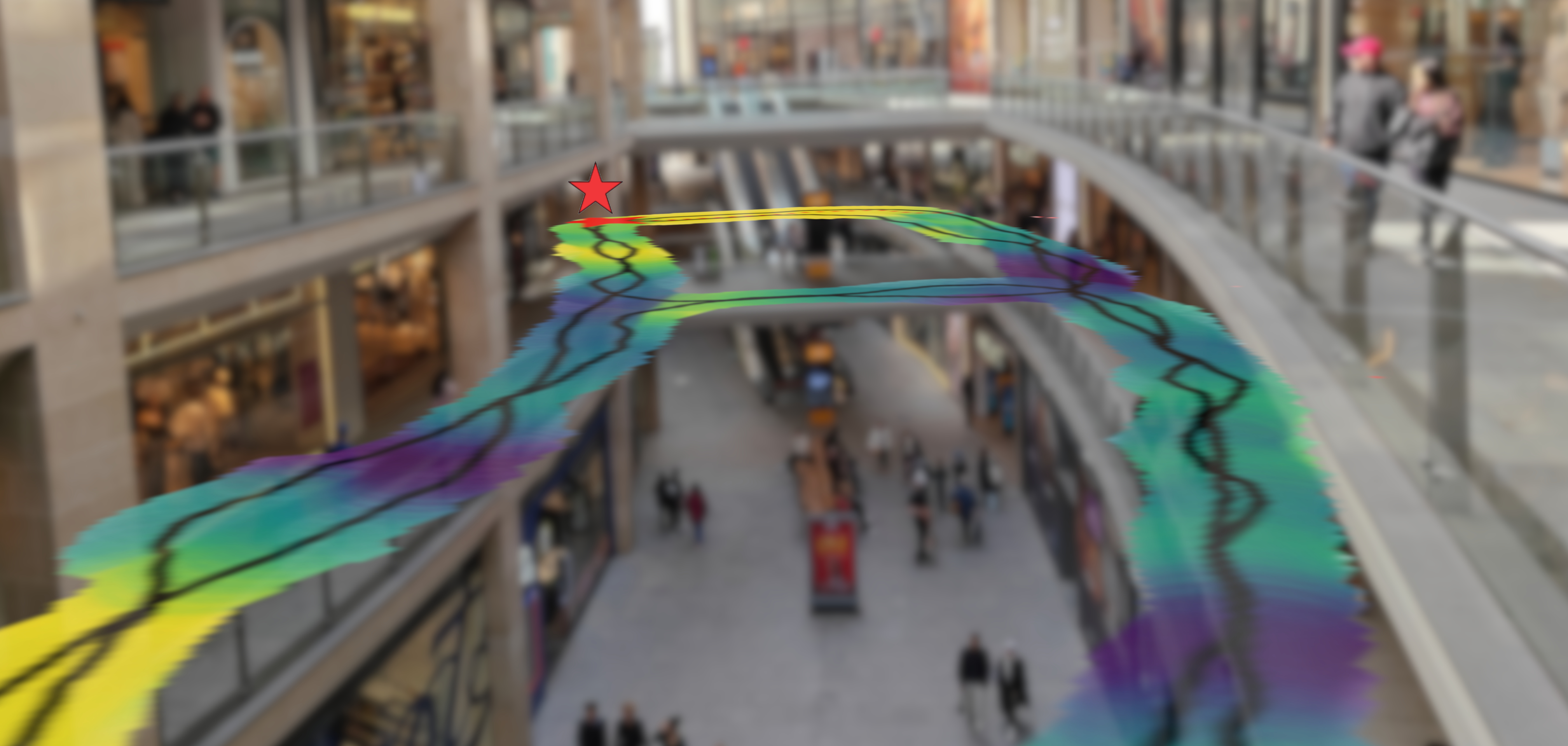}
\label{fig:st_james_overlay}}
\qquad
\subfloat[][Trajectory and MF map (\mut)]{
\includegraphics[width=0.5\columnwidth]{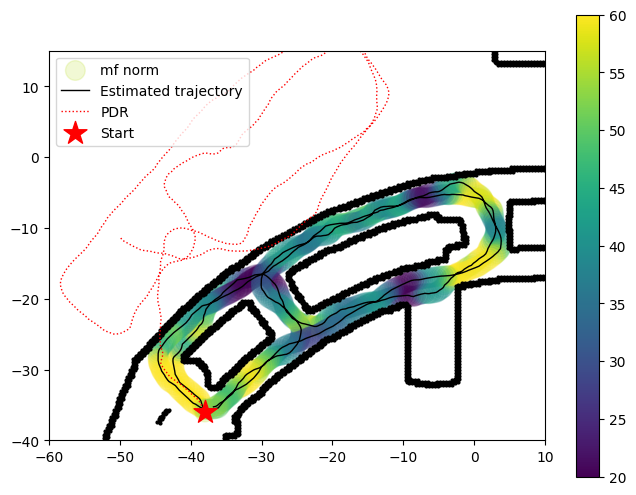}
\label{fig:calib_estimation_traj}}
\subfloat[][Calibration process]{
\includegraphics[width=0.5\columnwidth]{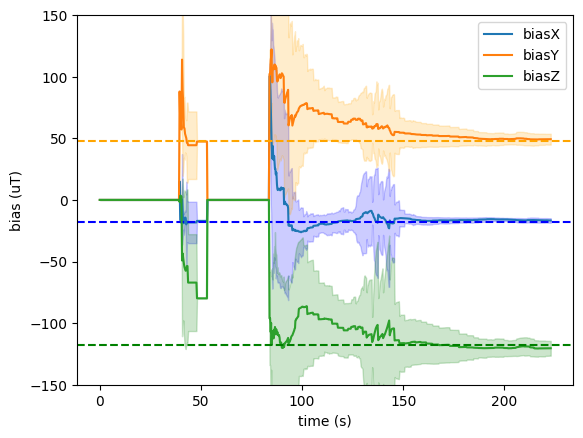}
\label{fig:calib_estimation_estimate}}
\qquad

\caption{(a-b) Estimated trajectory collected from a shopping mall and the produced MF map. (c) Corresponding bias estimation by particle-wise Kalman Filters. The dashed line is bias reported by the operating system. The band depicts three times the standard deviation of the bias over the particles. 
}
\label{fig:calib_estimation}
\end{figure}

One of the main ideas of particle filtering is to focus computational resources on the most promising search areas. This is exactly what we leverage here: We use this principle by simulating trajectories and maps (\textit{known poses}) with a PF. When a particle encounters a loop closure (partial sphere surface), we update its KF bias estimate. This is done in the current (particle-wise) sensor frame by finding the bias change (residual) that minimizes the difference between the current measurement and map estimate. This residual is used directly in the Kalman Filter. This process, illustrated in Figure \ref{fig:calib_estimation}, effectively guides computation toward the most likely maps and calibrations. Incorrect biases and maps will produce essentially random, inconsistent measurements and low weights \cite{robertson2013simultaneous}.

\subsection{Simplifying assumptions}
For the purposes of this paper, we make several simplifying choices and assumptions: We  estimate only the bias offset vector $\mb{b} \in \mathbb{R}^3$ instead of more general calibration \cite{kok2016magnetometer} and assume the bias stays constant during the experiment. We use trajectories with several overlapping segments, and assume that the magnetometer stays at approximately constant height and that its yaw is aligned with the movement direction.

\section{Simultaneous Localization, Mapping, and Calibration (SLAMnC)}
\subsection{Rao-Blackwellized particle filter}

To implement the intuition described in Section \ref{section:intuition}, we use a PF factorization commonly known as the Rao-Blackwellised Particle Filter (RBPF) \cite{stachniss2007analyzing}. It estimates the joint posterior $p(X_t, m_t, c_t | Z_t, U_t)$ of the trajectory of poses $X_t$, the map $m_t$, and calibration $c_t$, conditioned on the observations $Z_t$ and controls $U_t$. Assuming that the calibration and map are independent of the control inputs given known poses and measurements, leads to the following factorization:

\begin{align}
p(X_t, m_t, c_t &| Z_t, U_t) \notag \\	
 						=&\underbrace{p(c_t | m_t)}_\text{calib $\approx$ KF} \underbrace{p(m_t | X_t, Z_t)}_\text{map} \underbrace{p(X_t | Z_t, U_t)}_{\approx \Particles}.
\end{align}

The right side is recursively approximated by a PF. The calibration is estimated using a KF, similarly to \cite{siebler2023magnetic_robot}, and the map for each particle is built analytically from the poses and measurements. That is, each particle $\xIT \in \Particles$ carries its own calibration $c_{t}^{(i)}$ and map $\mIT$. The measurement weight update for the PF is given by
\begin{align}
	\wIT	&= p(\z_t | \xIT, c_{t-1}^{(i)}, \mITPrev)\wITPrev. \label{formula:weigh_update}
\end{align}
Calibration $c_{t}^{(i)}$ for each particle is updated using a Kalman Filter, followed by updating maps $\mIT$ analytically based on the trajectory $X_t^{(i)}$ and the measurements $Z_t$. For simplicity, the calibration estimated in this paper refers only to the bias vector $\mb{b} \in \mathbb{R}^3$, and we use bias and calibration interchangeably.

\subsection{Map representation and MF estimates}
Our aim is to align the magnetic field (MF) map estimate with the current sensor frame, enabling computation of the residual for the current bias. Each data point $d_i = (\z_t, \mb{x_i}, \theta_i, r^{sp}_t, t)$ in our map includes the measurement $z_t$ in the sensor frame, 2D coordinates $\mb{x}_i$, heading $\theta_i$, rotation $r^{sp}_t(\z)$ from sensor to planar frame, and a timestamp $t$. We refer to these elements using dot notation; for example, $d_i.\z$ refers to the measurements associated with the data point $d_i$. Note that $\z_t$ and $r^{sp}_t$ are common to all data points (at time $t$), while $\mb{x}_i$ and $\theta_i$ depend on the estimated trajectory and vary for each map (particle). We define $r^{sw}_t(\z)$ as the rotation from the sensor to the world frame, i.e., $r^{sp}$ followed by rotation by $\theta_i$ around the $z$ axis.

Assuming we know the map $m$ and the bias vector $\mb{b} \in \mathbb{R}^3$, we can estimate the MF in the world frame at a given location using the map estimate function:
\begin{align}
m^w(x, \mb{b}) = f(x, r^{bsw}(n(x), \mb{b})).
\end{align}

\begin{figure}[ht]
\centering
\includegraphics[width=1.0\columnwidth]{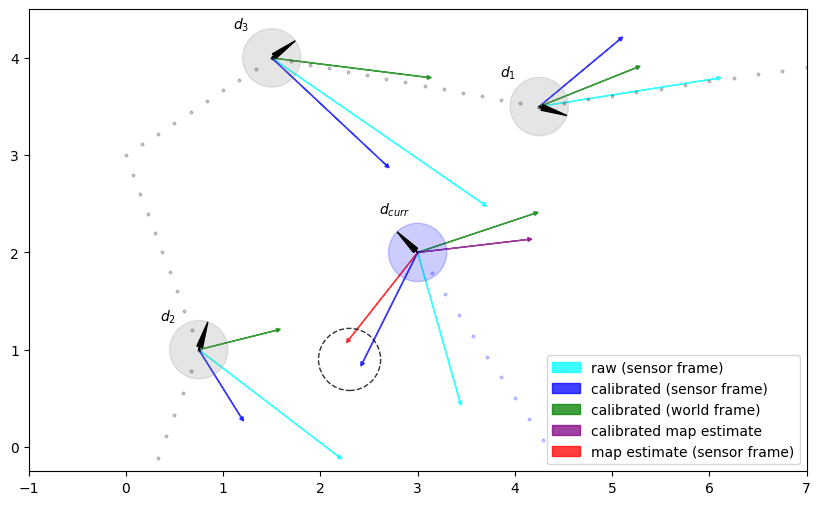}
\caption{2D map estimate construction at $d_{\text{curr}}$ from data points $d_1, d_2, d_3$. First, the calibration is applied to all data points. Then the readings are rotated to the world frame, in which the map estimate is computed (in this example the mean of kNN). Finally, the map estimate is rotated to $d_{\text{curr}}$ sensor frame, where it is used to obtain the residual for the PF and KF (vector difference inside the dashed circle).}
\label{fig:map_estimate}
\end{figure}

Here $n(x)$ represents the spatial neighbors of $x$, such as kNN or those within a certain radius. The function 

\begin{align}
r^{bsw}(d, \mb{b}) = r^{sw}_t(d.z - \mb{b})
\end{align}

removes the bias followed by rotation from the sensor to the world frame. Function $f(x, \mb{n}^w)$ is a MF estimate (e.g. interpolation or given by a GP \cite{solin2016terrain, viset2022extended}) function at $x$ based on the world-coordinate neighborhood $\mb{n}^w$. For simplicity in notation, these functions return corresponding mapped sets when applied to sets rather than single values. Finally, we rotate the map estimate to the current sensor frame. This MF estimate in the current (particle-wise) sensor frame is denoted by $m(x, \mb{b})$ and is used to compute residuals needed for the PF and KF. Figure \ref{fig:map_estimate} illustrates this process with a 2D example. 

\subsection{Kalman Filter for bias estimation}
In addition to the MF map, each particle $k$ carries a Kalman Filter (KF) estimating the bias vector $\mb{b} \in \mathbb{R}^3$. We use a standard Kalman Filter, similar to \cite{siebler2023magnetic_robot}. The units in the following are microteslas (\mut), and we treat each channel independently. With no prior information about the initial state, we set $\mb{u}_0 = [0, 0, 0]^T$ at the origin and use an inflated initial covariance $\mb{P}_0=\diag(1000^2, 1000^2, 1000^2)$. To avoid modeling potential measurement errors, we also use high values for measurement noise $\mb{R}=\diag(200^2, 200^2, 200^2)$. The process noise $\mb{Q}=\diag(2^2, 2^2, 2^2)$ is kept relatively low to ensure stability post-convergence. 

The KF "measurement" for each particle is derived from the particle's previous map estimate and previous bias rotated to the current sensor frame: $m^{(k)}_{t-1} (x^{(k)}, \mb{b}^{(k)}_{t-1})$. In the KF update step, it yields us a residual
\begin{align}
y^{(k)}_{t} = (z_t - \mb{b}^{(k)}_{t-1}) - m^{(k)}_{t-1} (x^{(k)}, \mb{b}^{(k)}_{t-1})
\label{formula:residual}
\end{align}
Although this “measurement” can be particularly noisy before the KF converges to accurate bias values, our experiments indicate that it provides a useful approximate residual for the actual bias. Figure \ref{fig:map_estimate} visualizes the computation of this residual in a 2D context.

\subsection{Likelihood functions for the particle filter}
For the MF likelihood in (\ref{formula:weigh_update}), we use both narrow and wide likelihood functions: $\exp{(-\dhat(y^{(k)}_{t})^2/(2\sigma^2))}$, where $\sigma \in \{10.0, 50.0\}$ and $\dhat$ clamps the residual (\ref{formula:residual}) between measurement and map estimate within two standard deviations. The narrow likelihood is used when the map estimate comes from spatially close points, i.e., when the estimate is less uncertain. Clamping protects against outliers, for example. In the shopping mall experiment, we also use a simple geometric constraint based on the floor plan, similar to map matching \cite{kaiser2011human}. We slightly penalize particles close to walls and give a very harsh penalty to particles intersecting with a wall. 

\subsection{Spatial neighborhood and weight update conditions}
When working with sparse MF maps, it is often beneficial to ignore or assign higher uncertainty to measurements (data points) in the map that are (a) temporally too close, or (b) spatially too far \cite{viset2022extended, lee2020magslam, gao2017signal, vallivaara2011magnetic}. Temporally close measurements are highly correlated with the current measurement and, when used without care, can distort the local estimate. Measurements that are spatially too far contain little to no information about the local MF. In fact, using them might give severely biased local estimates. To mitigate (a), neighborhoods exclude points that are temporally closer than \SI{5.0}{s} from the current measurement. To address (b), we define conditions on the closeness of the neighborhood to determine how or if we want to update our PF and/or KF. Because faraway MF measurements do not produce reliable information, we update the KF only if the closest point in the neighborhood is within \SI{0.25}{m}. This is also the distance condition for our tighter PF likelihood ($\sigma = 10$). The distance condition for the wider likelihood ($\sigma = 50$) is \SI{2.0}{m}, which is primarily for coarse heading information. The PF weights are not updated for particles that have not yet updated their bias estimate. 

\subsection{Practical implementation and considerations}
The ideas presented above can be implemented in various ways, and we expect similar performance despite minor differences in implementation details (e.g., MF map estimate, motion model, map representation). In our experiments, we use machinery developed earlier for compact map representation \cite{eliazar2003dp, vallivaara2018quadtree} and managing multiple conditioned likelihood functions \cite{vallivaara2013monty}. 
Pseudocode in Algorithm \ref{alg:slamc} outlines a naive, high-level view of the algorithm that omits many details, such as how to handle measurements in non-mapped areas. For these specifics, we refer to previous work \cite{lee2020magslam, vallivaara2013monty, vallivaara2018quadtree}. 

\begin{algorithm}[ht]
\scriptsize
\DontPrintSemicolon
\caption{Simultaneous Localization, Mapping and Calibration (SLAMnC)}
\BlankLine
\label{alg:slamc}
\KwIn{ 
$\ParticlesPrev$: particles, $\WeightsPrev$: weights, $\z_t$: measurement, $\u_t$: control
}
\KwOut{
$\Particles$: updated particles, $\Weights$: updated weights
}
\BlankLine
$\ParticlesOver \gets \emptyset$, $P \gets |\ParticlesPrev|$ \\
\nlset{Motion model}\For{$i \gets 1$ \KwTo $P$}{ \label{alg:pf_predict}
draw $\xIT$ from $\motionModelI$ \\
$\ParticlesOver \gets \ParticlesOver \cup \xIT$ 
}
\BlankLine
$\Weights \gets \emptyset$ \\ 
\nlset{Measurement update}\For{$i \gets 1$ \KwTo $P$}{ \label{alg:pf_measurement_update}
$\xIT$.computeSpatialNeighborhood() \\
\nlset{KF update}\If{$\xIT$.hasCloseMfNeighbors}{
\nlset{Residual}$y^{(k)}_{t} = (\z_t - \mb{b}^{(k)}_{t-1}) - m^{(k)}_{t-1} (\x^{(k)}, \mb{b}^{(k)}_{t-1})$ \\
$\xIT$.KF.update($y^{(k)}_{t}$)  \\
}
$w \gets 1$ \\
\nlset{MF weight update}\If{$\xIT$.hasBiasEstimate}{
$w \gets p(\z_t | \xIT, c_{t-1}^{(i)}, \mITPrev) \cdot w $  \\

}
\nlset{Floor plan weight update}\If{floorPlanAvailable}{
$w \gets p(\text{floorPlan} | \x_t^{(i)}) \cdot w $\\
}
$\wIT \gets w \cdot \wITPrev$ \\
$\Weights \gets \Weights \cup \wIT$ \\

\nlset{MF map update}  $\xIT$.updateMap($\z_t$) \\
}
$\Weights \gets$ normalize($\Weights$) to sum to $1$
\BlankLine
$\Particles \gets \text{resampleIfNeeded}(\Particles)$ \\
	\BlankLine
\Return{$\Particles$, $\Weights$}
\end{algorithm}

\section{Experiments}\label{section:experiments}
 We conduct two experiments to validate our approach. The first experiment uses mobile phone data from a shopping mall, characterized by high PDR and orientation uncertainty. The second experiment uses mobile robotics data from indoor environments, where pitch and roll are fixed, and odometry is much more reliable. The first experiment uses geometric constraints in addition to MF and PDR. The second experiment uses only MF and odometry. Despite the distinct dynamics, we apply our method identically for both scenarios. Apart from different step lengths, the only variation is in the motion model parameters. The data used in our experiments are spatially downsampled to \SI{0.2}{m} and \SI{0.1}{m} steps for phone and robot, respectively, with corresponding MF and orientation measurements. The particle count is $5000$.

\subsection{Nearest Neighbor Error (NNE) and compass heading} \label{section:nne}
Due to the lack of accurate ground truth, we propose a simple error metric to estimate MF map consistency. This metric is the difference in world coordinate MF vectors between each measurement and its spatially nearest neighbor, with some exclusions to include neighbors only from different parts of the trajectory: (a) the neighbor has to be temporally at least \SI{5}{s} apart (excluding itself), and (b) spatially closer than \SI{7}{m}. In our experience, despite its simplicity, high NNE correlates strongly with PF divergence. We report our results as the median norm of the NNE error over all eligible data points in the map. Additionally, given the importance of heading in many applications \cite{khider2013characterization}, we briefly confirm that the compass heading in the world frame properly aligns with magnetic north, as it should with a well-calibrated sensor (Fig. \ref{fig:compass_heading}).

\section{Experiment: Smartphone data from a shopping mall} \label{section:experiment_phone}
\begin{figure}[ht]
\centering
\includegraphics[width=1.0\columnwidth]{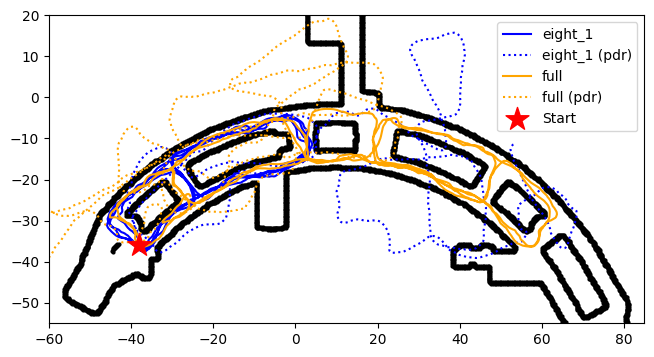}
\caption{St James Quarter shopping mall in Edinburgh. Trajectories \text{eight\_1} and \text{full} from our shopping mall data set. Trajectory \text{eight\_2} is visualised in \ref{fig:calib_estimation_traj}.}
\label{fig:shopping_mall}
\end{figure}
The main goal of this experiment is to verify that our method can estimate bias and produce results comparable to the operating system's bias estimate (OS-bias). We compare the absolute bias estimation values and the consistency of the resulting MF map.

The data set for this experiment was collected from St James Quarter shopping mall in Edinburgh during business hours with medium crowding (Fig. \ref{fig:st_james_overlay}). We used the Sensor Logger Android application (\url{https://www.tszheichoi.com/sensorlogger}) on a OnePlus 11 smartphone. The raw sensor data  was processed to PDR output using a modified local fork of RoNIN ResNet \cite{herath2020ronin} with a pre-trained model (ronin\_resnet.zip). The trajectories in the data set do not contain calibration movements in a homogeneous field, as shown in Fig. \ref{fig:rotation_and_calib}. The MF data that we use is from the uncalibrated magnetometer sensor. We use the acceleration to obtain pitch and roll for the rotation to the planar coordinate frame. We estimate the OS-bias ($[-18, 48, -118]^T$) by collecting a separate calibration movement trajectory and extracting the sphere's center (Fig. \ref{fig:calib_movement_calib}). Spatial PDR downsampling further ensures we do not rotate the phone in a constant field.

Often the agent movement is constrained by the physical environment, such as corridors. Many magnetic field SLAM papers limit themselves to very simple and constrained trajectories with overlapping and parallel segments \cite{solin2016terrain, kok2018scalable, viset2022extended, pavlasek2023magnetic}. This paper's shopping mall experiment is no exception: Two of our trajectories are simple figure-eight-shaped paths, ensured to contain data in multiple directions for each mapped corridor. One of our trajectories is a longer, floor-scale path. Although overlapping trajectories are a justifiable assumption and are even explicitly used as the basis of some algorithms \cite{robertson2013simultaneous}, we want to stress that this makes the problem considerably easier than unconstrained movement. We study the unconstrained case in the mobile robot experiment.

The trajectories we analyze here are \text{eight\_1} (longer simple trajectory), \text{eight\_2} (shorter simple trajectory with deliberate pitch change), and \text{full} (longer trajectory, spanning almost the whole floor). These are visualized in Figures \ref{fig:calib_estimation_traj} and \ref{fig:shopping_mall}. In addition to evaluating the bias estimation, we run SLAM on the trajectories with four different configurations. Our primary focus is on the configuration with uncalibrated (raw) magnetometer data and bias estimation (\text{SLAMnC, raw}). The next two use OS-calibration, with (\text{SLAMnC, os\_calib}) or without (\text{SLAM, os\_calib}) running bias estimation on top. The final one (\text{no\_mf, os\_calib}) does not use the MF data in SLAM at all. This configuration is included to evaluate how much the MF contributes in addition to PDR and geometric constraints. We compute the Nearest Neighbor Error (NNE) as defined in Subsection \ref{section:nne} and compass headings to evaluate the resulting MF map consistency.

\subsection{Results}
\begin{figure}[ht]
\centering
\subfloat{
\includegraphics[width=1.0\columnwidth]{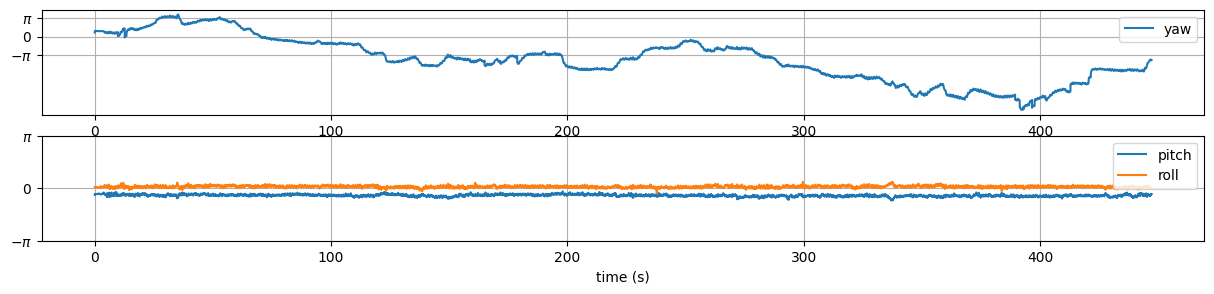}
}
\qquad
\subfloat{
\includegraphics[width=1.0\columnwidth]{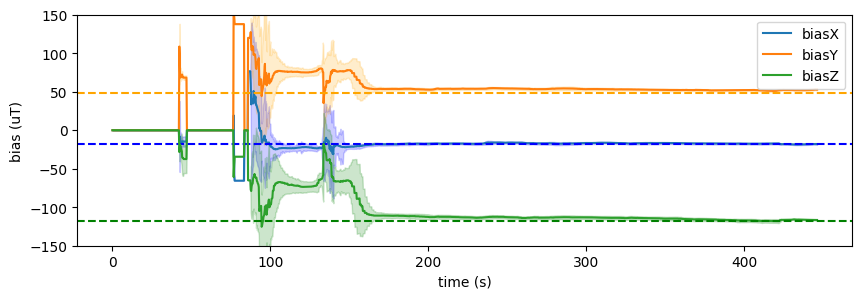}}
\qquad
\subfloat{
\includegraphics[width=1.0\columnwidth]{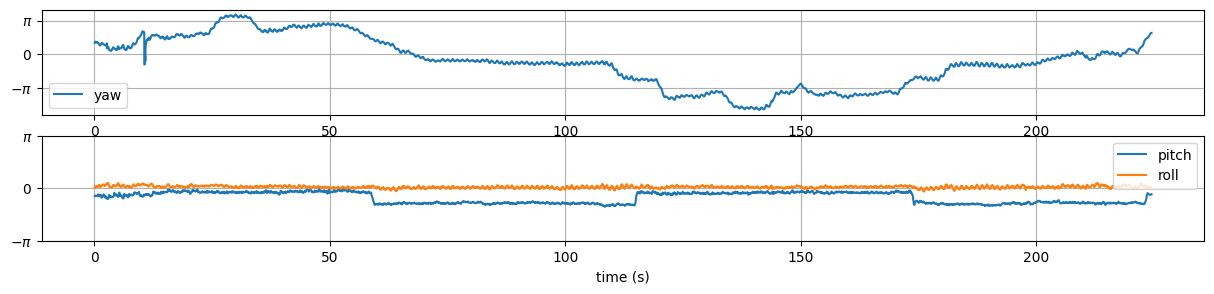}
}
\qquad
\subfloat{
\includegraphics[width=1.0\columnwidth]{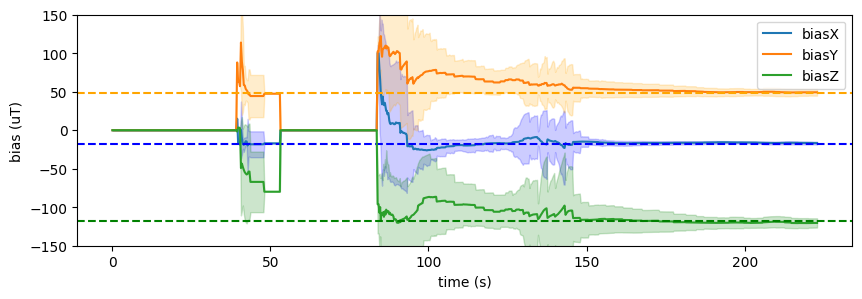}
}
\qquad
\caption{Phone orientation and bias estimation process for shopping mall trajectories eight\_1 (top two) and eight\_2 (bottom two). Even with very small amount of non-yaw rotations, the bias estimate converges very close to OS-bias (dashed line).}
\label{fig:rotation_and_calib}
\end{figure}

In all of our three test trajectories, our method converges very close to the OS-bias
 without any calibration movements beyond walking with the phone in hand (Fig. \ref{fig:rotation_and_calib}). The typical difference is less than 5\mut \ per axis. The convergence time varies between trajectories and axes (\SI{90}{s}--\SI{170}{s}), and is highly dependent on loop closures, as expected.

\begin{figure}[ht]
\centering
\includegraphics[width=1.0\columnwidth]{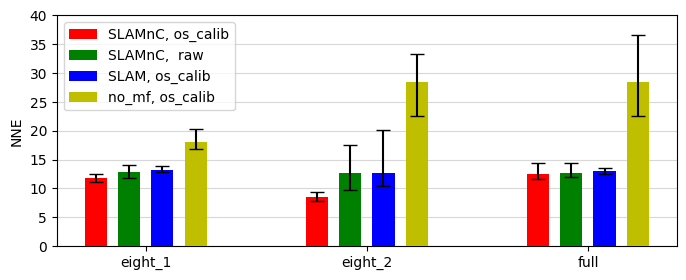}
\caption{Nearest Neighbor Error on three shopping mall trajectories over four configurations. Results are averages over five runs and the error bar depicts the range.} 
\label{fig:map_errors_mall}
\end{figure}

\begin{figure}[t!]
\centering
\subfloat[Subfigure 1 list of figures text][Phone compass headings in St James Quarter (full)]{
\includegraphics[width=1.0\columnwidth]{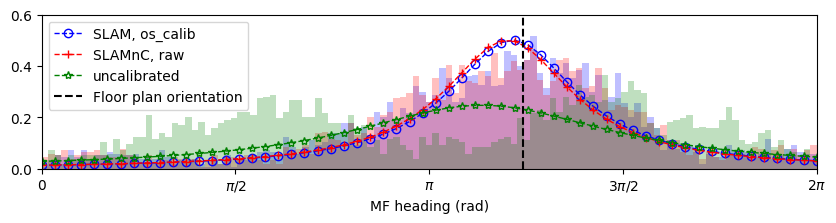}
\label{fig:compass_heading_phone}}
\qquad
\subfloat[Subfigure 3 list of figures text][Robot compass headings in Apartment]{
\includegraphics[width=1.0\columnwidth]{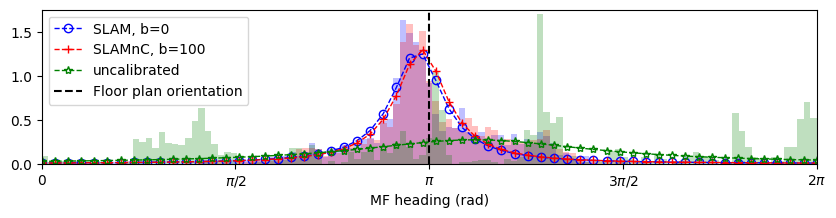}
\label{fig:compass_heading_robot}}
\qquad
\caption{Estimated compass headings from the shopping mall (phone) and apartment (robot) experiments. Cauchy distributions fitted to the histogram data are almost identical for SLAMnC and pre-calibrated data, i.e. we are able to extract very similar compass headings.}
\label{fig:compass_heading}
\end{figure}

The NNE results (Fig. \ref{fig:map_errors_mall}) show that our method with uncalibrated data (\text{SLAMnC, raw}) produces almost identical output to using the OS-bias, and slightly improves over OS-bias when using it as a starting point (\text{SLAMnC, os\_calib}). We note that the OS might not provide perfect bias either. Not using the MF for positioning leads to inconsistent MF maps; in \text{eight\_1} and \text{full}, this is explained by the PF consistently diverging despite the geometric constraints. The compass headings from SLAMnC are almost identical to those using the OS-bias (Fig. \ref{fig:compass_heading_phone}).

\section{Experiment: Mobile robot data from office and apartment environments} 
\label{section:experiment_robot}
The main goal of this experiment is to verify that our approach is applicable to mobile robotics and can estimate bias regardless of its magnitude. We use mobile robot MF data collected from office and apartment-style environments in Oulu, Finland \cite{vallivaara2011magnetic} (\url{https://github.com/scilari/mf_slam_data}). One trajectory is from a Computer Science and Engineering lobby (CSE lobby, Fig. \ref{fig:cse_lobby}) and one from a private apartment (Apartment, Fig. \ref{fig:apartment}). Since the odometry is much more accurate than the PDR in the previous experiment, no geometric constraints are used here. Thus, we only use the odometry and magnetometer readings for SLAM.  

If the magnetometer is centered on a differential wheel robot \cite{siebler2023magnetic_robot, vallivaara2011magnetic}, obtaining the bias in the horizontal plane by simply rotating in place would be trivial. To avoid this, we use the same temporal constraint as in the mall experiments, i.e., not using measurements temporally closer than \SI{5}{s}. This makes our experiment correspond to a robot driving forward, similar to \cite{siebler2023magnetic_robot} (Fig. \ref{fig:siebler}). Estimating $z$-axis bias does not make much sense in this scenario (constant pitch and roll). However, we keep it in the estimator for consistency and to verify that it behaves as expected, i.e., stays close to zero.

For this experiment, we add bias of varying magnitudes ($5, 10, 50, 100 \pm 10\%$ per $x, y$ component) to our raw sensor readings and compare the PF performance with and without our bias estimation.

\subsection{Results}
\begin{figure}[ht]
\centering
\subfloat[][CSE lobby]{
\includegraphics[width=0.5\columnwidth]{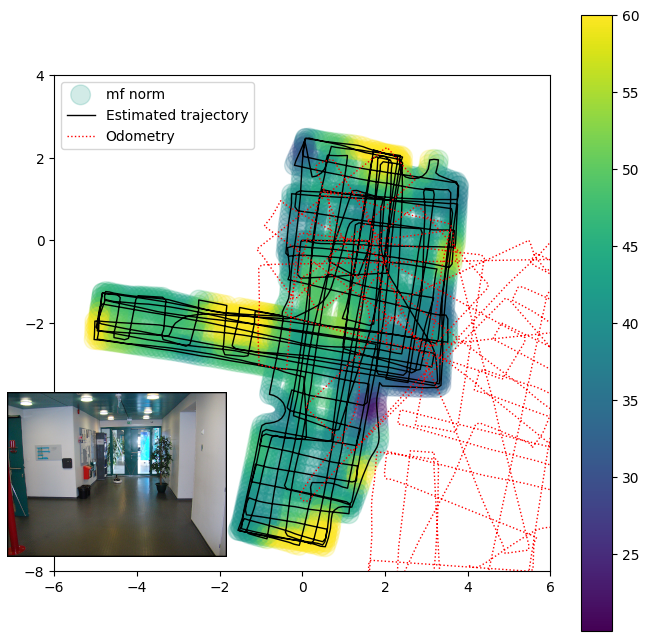}
\label{fig:cse_lobby}}
\subfloat[Subfigure 3 list of figures text][Apartment]{
\includegraphics[width=0.5\columnwidth]{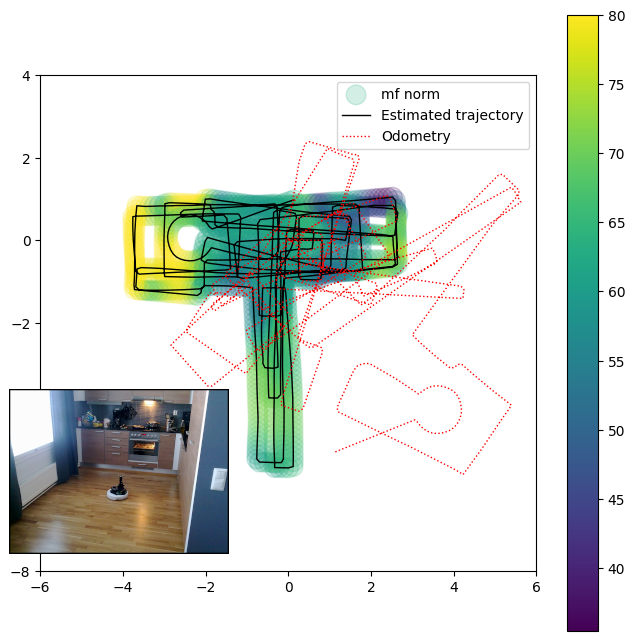}
\label{fig:apartment}
}
\qquad
\subfloat[][\text{Bias estimation}]{
\includegraphics[width=0.5\columnwidth]{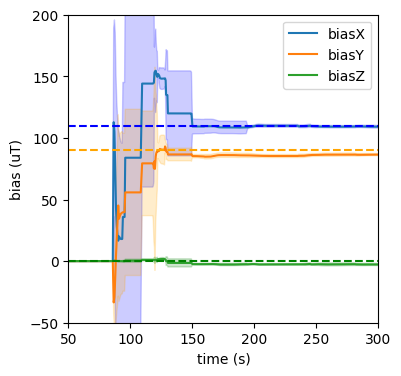}
}
\subfloat[][Bias estimation]{
\includegraphics[width=0.5\columnwidth]{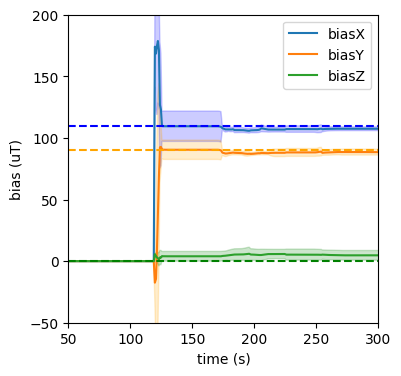}
}
\qquad
\subfloat[][NNE comparison]{
\includegraphics[width=0.5\columnwidth]{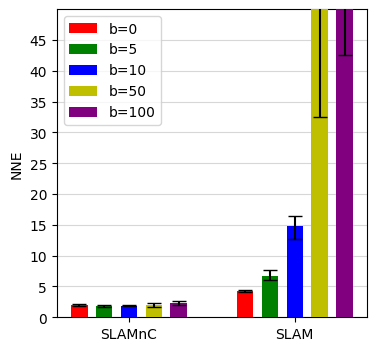}
\label{fig:cse_lobby_nne}
}
\subfloat[][NNE comparison]{
\includegraphics[width=0.5\columnwidth]{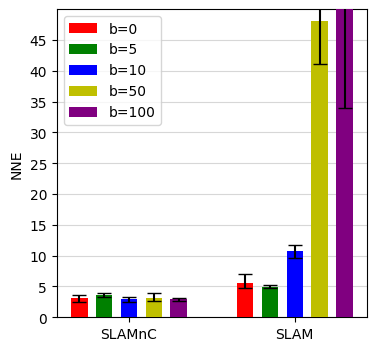}
\label{fig:apartment_nne}
}
\qquad
\caption{Mobile robotics experiments in an office (a) and apartment (b) environments. SLAMnC is able to consistently estimate the bias (middle row) and produce consistent MF maps regardless of bias. This is clearly reflected in the Nearest Neighbor Error (bottom row). The NNE results are averages over five runs and the error bar depicts the range.}
\label{fig:robot_experiment}
\end{figure}

Our method achieves a 100\% success rate with all bias magnitudes: it estimates the bias correctly and converges to maps that visually match the environment geometry (Fig. \ref{fig:robot_experiment}). When bias estimation is turned off, the PF starts to diverge at a bias magnitude of around 10 (\mut). This is clearly reflected in the MF map consistency: with our method, NNE stays constant, whereas without bias estimation, the error rapidly increases with bias magnitude (Fig. \ref{fig:cse_lobby_nne} and \ref{fig:apartment_nne}). The bias estimation plot is cut off at $t=300$ to show the convergence region (the estimate is practically constant thereafter). Again, \text{SLAMnC} slightly improves over SLAM with manual calibration (\text{SLAM, b=0}), suggesting that there may still be some minor (below 2 \mut) bias left even after the manual calibration. The compass headings estimated by \text{SLAMnC} are almost identical to manually calibrated data (Fig. \ref{fig:compass_heading_robot}).

\section{Limitations, future work, and conclusions} \label{section:limitations}
The presented approach has several limitations, some of which we believe are fundamental (F) and others that should be solvable with more research (R) or different system design (S): 
\textbf{Constant bias} assumption (F) - this might not be generally true, e.g., due to magnetization; 
\textbf{Long trajectory} requirement (F) - multiple loop closures are needed, although incremental construction (R/S) in a crowd-sourcing manner could be possible \cite{wang2023crowdmagmap};
\textbf{Fixed phone orientation to walking direction} (R) -  this can be solved by good orientation estimates \cite{herath2020ronin};
\textbf{Planar robots don't provide $z$ bias} (S/R) - this might be a non-issue, and also fixable with points of reference for the $z$ values (e.g., incremental map), accurate compass heading alone can also be very useful;
\textbf{Bias offset only calibration} (R) - more general calibrations can be definitely estimated.

In future work, we plan to study the calibration search space and residuals. Our current approach with a KF is rudimentary and can likely be improved by proper analysis and the introduction of more sophisticated filters (e.g., UKF or EKF \cite{kok2016magnetometer}). Informed bias initialization could improve the filter convergence, and we could take measures to escape from a divergent state. The PF likelihood functions could be made adaptive based on the bias uncertainty. There is also the possibility of a crowd-sourcing-like approach, where previously collected map data is used as a starting point for SLAMnC. It could also be combined with SLAC to handle dynamic biases. Nothing limits us to using only MF data for position estimation; replacing it with visual SLAM, using radio-based positioning, or using movement likelihood assumptions as in \cite{robertson2013simultaneous} would be straightforward. Testing the proposed approach on more comprehensive data sets with proper ground truth is also warranted.

We have presented \text{SLAMnC}, an RBPF-based SLAM approach for bias estimation, that consistently estimates the magnetometer bias without needing explicit calibration movements. Despite its limitations and straightforward formulation, the preliminary results are very promising and provide a good starting point for further research to eliminate the need for manual calibration.

\bibliographystyle{elsarticle-num} 
\bibliography{bibliography}

\end{document}